%% file: main.tex
\pdfoutput=1
\documentclass{article}

\usepackage[preprint,nonatbib]{nips_2018}

\usepackage[utf8]{inputenc} 
\usepackage[T1]{fontenc}    
\usepackage{url}            
\usepackage{booktabs}       
\usepackage{amsfonts}       
\usepackage{nicefrac}       
\usepackage{microtype}      

\usepackage{bookmark}

\usepackage{color}
\usepackage{xcolor}
\usepackage{caption}
\usepackage{wrapfig}
\usepackage{subfig}

\usepackage[]{algorithmicx}
\usepackage[ruled]{algorithm2e}

\usepackage{graphicx}

\usepackage{amsbsy}
\usepackage{amsmath}
\usepackage{amssymb}
\usepackage{multirow}
\usepackage{array,makecell}
\usepackage{tabularx}

\newcommand{\snet}{ScreenerNet}

\newcommand{\etal}{\emph{et al.}}

\makeatletter
\def\BState{\State\hskip-\ALG@thistlm}
\makeatother

\newcommand{\titleval}{ScreenerNet: Learning Self-Paced Curriculum for Deep Neural Networks}

\title{\titleval}

\author{
  Tae-Hoon Kim
  \\
  Intel Corporation\\
  Seoul, Korea\\
  \texttt{pete.kim@intel.com}\\
  \And
  Jonghyun Choi \\
  Allen Institute for Artificial Intelligence\\
  Seattle, WA 98103 \\
  \texttt{jonghyunc@allenai.org} \\
}

\begin{document}

\maketitle

\begin{abstract}
\input{abstract.tex}
\end{abstract}

\section{Introduction}
\input{intro.tex}

\section{Related Work}
\input{related.tex}

\section{Approach}
\input{approach.tex}

\section{Experiments}
\input{exp.tex}

\section{Conclusion}
\input{conclusion.tex}








\input{main.bbl}
\bibliographystyle{plain}

\end{document}

%% file: abstract.tex
We propose to learn a curriculum or a syllabus for deep reinforcement learning and supervised learning with deep neural networks by an attachable deep neural network, called \emph{\snet}.
Specifically, we learn a weight for each sample by jointly training the \snet\ and the main network in an end-to-end self-paced fashion. 
The \snet\ has neither sampling bias nor memory for the past learning history.
We show the networks augmented with the \snet\ converge faster with better accuracy than the state-of-the-art curricular learning methods in extensive experiments of a Cart-pole task using Deep Q-learning and supervised visual recognition task using three vision datasets such as Pascal VOC2012, CIFAR10, and MNIST. Moreover, the \snet\ can be combined with other curriculum learning methods such as Prioritized Experience Replay (PER) for further accuracy improvement.

%% file: intro.tex
Training a machine learning model with chosen training samples in a certain order improves the speed of learning and is called \emph{Curriculum Learning}~\cite{bengioWG10}.
The curriculum learning recently gains much attention due to the difficulty of training deep models for reinforcement learning~\cite{Andrychowicz2017,Graves2017,Sukhbaatar2017}.
However, selecting and ordering samples is a hard-decision process and significantly lowers the chance of samples to be selected in the later iterations and changes solution to which the model converges if the samples are rejected earlier (sampling bias).
In addition, the decision criteria are mostly defined by a set of hand-crafted rules such as classification error or confidence of the main network in majority of previous work~\cite{Bengio2015,bengioLCW09,Graves2017}.
Those hand-crafted rules require additional rules to handle the sampling bias.

\begin{wrapfigure}{r}{.5\textwidth}
    \vspace{-1.5em}
    \centering
    \includegraphics[width=0.98\linewidth,bb=4 3 705 287]{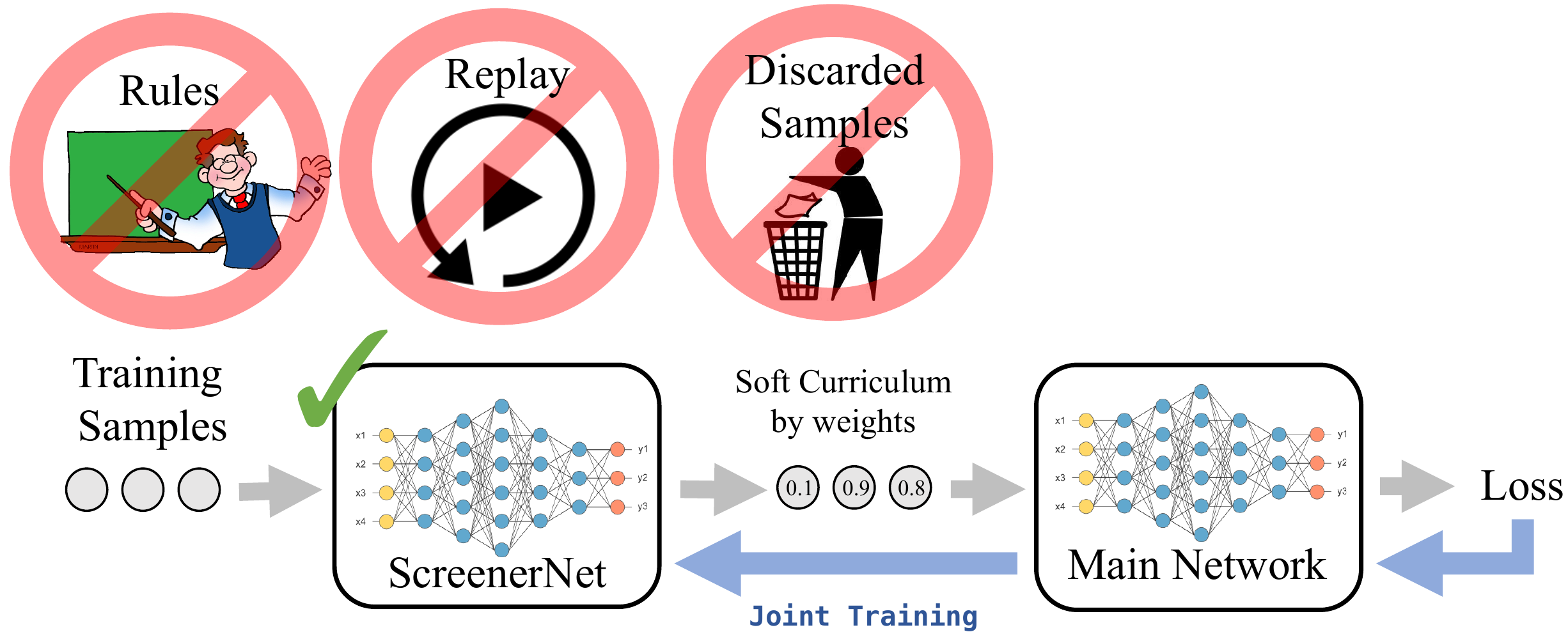}
    \caption{\snet\ learns the valuation of training samples for a main network. It regresses the sample into its significance for training; significant if it is hard for the current performance of the main network, and non-significant if it is easy.}
    \label{fig:teaser}
    \vspace{-.5em}
\end{wrapfigure}

To alleviate the sampling bias, we present a scheme to determine a soft decision by a weight value of every training sample for building a curriculum. Moreover the soft decision is learned in a self-paced manner~\cite{kumarPK10,jiangMZSH15,jiangMYLSH14,supancicR13,maMXLD17} without requiring to memorize the historic loss values~\cite{Schaul2016}. This is a generalization of the hard decision by considering all samples at every curriculum update, thus never ignores any samples at any iteration to maximize the efficacy of the curricular learning, motivated from the recent work of efficiently exploring spaces of state, action, and goal through the learning~\cite{Andrychowicz2017, Sukhbaatar2017,jiangMZSH15} and is similar to~\cite{Hinton2007,Schaul2016,jiangMZSH15}.
Moreover, to discover the rules for curriculum at each training iteration that are beyond our intuition, we learn the curriculum by an attachable neural network along with the main network being trained.
We call the attachable network as \emph{\snet}.
The \snet\ is jointly trained with the main network in an end-to-end fashion as illustrated in Figure~\ref{fig:teaser}. 
The \snet\ provides the locally most accurate weights per the main network being trained in a self-paced manner.

Unlike the many recent approaches that are mostly applied to deep reinforcement learning problems~\cite{Schaul2016,Andrychowicz2017,fanTQBL17}, we empirically show that our \snet\ can also be applied to various types of networks including convolutional neural network for visual recognition as well as deep reinforcement learning.
The \snet\ not only expedites the convergence of the main network but also improves the accuracy at the end of the training.
In particular, in a real world application \emph{e.g.}, a multi-camera surveillance system, pre-training all target objects is non trivial due to varying environments. The application strongly requires incremental updates of neural network models. The networks in such embedded systems are designed to be small for computational benefit and energy efficiency in the inference time. \snet\ should help in such use cases.
Finally, the \snet\ can extend the stochastic sampling based curriculum learning approaches for further improvement of final accuracy.

%% file: related.tex
As a pioneering work of curriculum learning, Hinton~\cite{Hinton2007} introduced an error-based non-uniform sampling scheme by importance sampling to significantly speed-up training of a network on a digit classification using MNIST dataset. The idea of error-based correction has been widely used as one of the most popular cues in the curriculum learning literature.
Since Bengio \etal~\cite{bengioLCW09} coin the term \emph{curriculum learning} for the method, number of approaches~\cite{Graves2017,leeG11,Schaul2016,linGGHD17,zhou2017} have been proposed to sample, weigh, or sort the training examples to improve accuracy and/or expedite the training.
These methods require to design a hand-crafted rule and/or to run the training procedure twice to obtain accurate estimate of sample valuation. To avoid second training, Jiang \etal\ propose to estimate the sample valuation at the same time of training a learning model~\cite{jiangMZSH15} by adding a term in the objective function. This is an extension of Kumar \etal's work on \emph{Self-Paced learning}~\cite{kumarPK10} in the curriculum learning literature. Our method also regresses the sample significance at the time of training.

Recently, Schaul \etal~\cite{Schaul2016} proposed a sampling scheme to increase the replay probability of training samples that have a high expected-learning-progress determined by an optimization loss.
They also proposed a weighted importance sampling method to address the bias of sampling, which is similar to our weighting scheme. 
In stochastic gradient-based optimization perspective, Loshchilov and Hutter~\cite{Loshchilov2016} also proposed to sample mini-batches non-uniformly, called AdaDelta~\cite{ADADELTA} and Adam~\cite{Kingma2015}.

Most recently, Graves \etal~\cite{Graves2017} proposed an automatic curriculum learning method for LSTM for the NLP application. They define a stochastic syllabus by a non-stationary multi-armed bandit algorithm of getting a reward signal from each training sample. They defined mappings from the rates of increase in prediction accuracy and network complexity to the reward signal. 

Koh and Liang~\cite{Koh2017} adopted influence functions from robust statistics to measure the effect of parameter changes or perturbations of the training data such as pixel values. It was applied to debugging models, detecting dataset error, and training-set attack. Although the direct application to curriculum learning is not presented, the authors show a way of predicting the significance of training samples, which can be useful for learning curriculum. 

Similar to our idea of attaching another network to the main one, self-play between the policy and a task-setting was proposed by Sukhbaatar \etal~\cite{Sukhbaatar2017}. 
The task-setter tries to find the simplest tasks that the policy cannot complete, which makes learning for the policy easy because the selected task is likely to be only just beyond the capability of the policy.
Similarly, Andrychowicz et al.~\cite{Andrychowicz2017} recently proposed to learn a curriculum but with a hindsight replay. They use unshaped reward signals even when they are sparse and binary, without requiring domain knowledge. 
These recent approaches address the sparsity and complexity of solution spaces in deep reinforcement learning, which is not a main issue for the supervised learning.

Very recently, Zhou \etal~\cite{zhou2017} proposed an adaptive feeding method that classifies an input sample as easy or hard one in order to forward the input to an appropriate one of a fast but less accurate neural network and an accurate but slow network for object detection. However, they only showed the speed up the inference preserving the accuracy of the main classifier.

In addition, Jiang \etal~\cite{jiangZLLL17} propose an attachable network to regularize the network not to overfit to the data with corrupted labels, called `MentorNet'. The MentorNet uses feature embedding of the sample by the main network, referred as StudentNet, and compute a weight for each sample. The MentorNet is very different from \snet\ as it is pretrained on other datasets than the dataset of interest for better regularization of corruptedly labeled samples whereas \snet\ aims for faster and better convergence of the training on the dataset of interest.


Unlike the previous approaches, our \snet\ has three benefits: First, it includes all samples to update weight even though the samples have weight values close to zero. 
This benefit is particularly significant in stochastic sampling approaches, if an important sample is assigned a low sampling priority at the early stage of training, it may not be likely to be picked again and may not be significantly used for the training until other samples have low sampling priority as well.
Second, it can learn a direct mapping of a training sample to its significance, even if the training sample is unseen, unlike the other memory-based methods.
%
Finally, it can extend the stochastic sampling curriculum learning approaches to improve further by taking the benefits of both methods.

%% file: approach.tex
We formulate the problem of building a curriculum by learning the importance of each sample for training the main network by a scalar weight in a joint learning framework.
Specifically, we define the importance of a training sample $\mathbf{x}$ at each iteration of training as a random variable $\mathbf{w}_\mathbf{x}$. 
Since we want to train the main network better in speed and accuracy, the objective of the curriculum learner maximizes the likelihood by sample weight $\mathbf{w}_\mathbf{x}$ given the error $\mathbf{E}$ between the prediction of the main network for $\mathbf{x}$ and its target. The maximum likelihood estimator of $\mathbf{w}_\mathbf{x}$ ($\widehat{\mathbf{w}_\mathbf{x}}$) can be written as follows:
\begin{equation}
\widehat{\mathbf{w}_\mathbf{x}} = \arg \max_{\mathbf{w}_\mathbf{x}} P(\mathbf{E}|\mathbf{w}_\mathbf{x}, \mathbf{W}_c)
							= \arg \max_{\mathbf{w}_\mathbf{x}} \frac{ P(\mathbf{w}_\mathbf{x} | \mathbf{E}, \mathbf{W}_c) P(\mathbf{E} | \mathbf{W}_c) }{ P(\mathbf{w}_\mathbf{x} | \mathbf{W}_c) }, 
\label{eq:wxi_org}
\end{equation}
where $\mathbf{W}_c$ is a parameter of the main network.
Since $\mathbf{E}$ is a function of $\mathbf{W}_c$, Eq.~\ref{eq:wxi_org} can be reduced to
\begin{equation}
\widehat{\mathbf{w}_\mathbf{x}} = \arg \max_{\mathbf{w}_\mathbf{x}} \frac{ P(\mathbf{w}_\mathbf{x} | \mathbf{E}) P(\mathbf{E} | \mathbf{W}_c) } { P(\mathbf{w}_\mathbf{x} | \mathbf{W}_c) }.
\label{eq:wxi}
\end{equation}
The \snet\ is a neural network to optimize the objective.

\subsection{\snet\ }

\begin{wrapfigure}{r}{.5\textwidth}
    \centering
    \vspace{-2em}
    \includegraphics[width=0.75\linewidth]{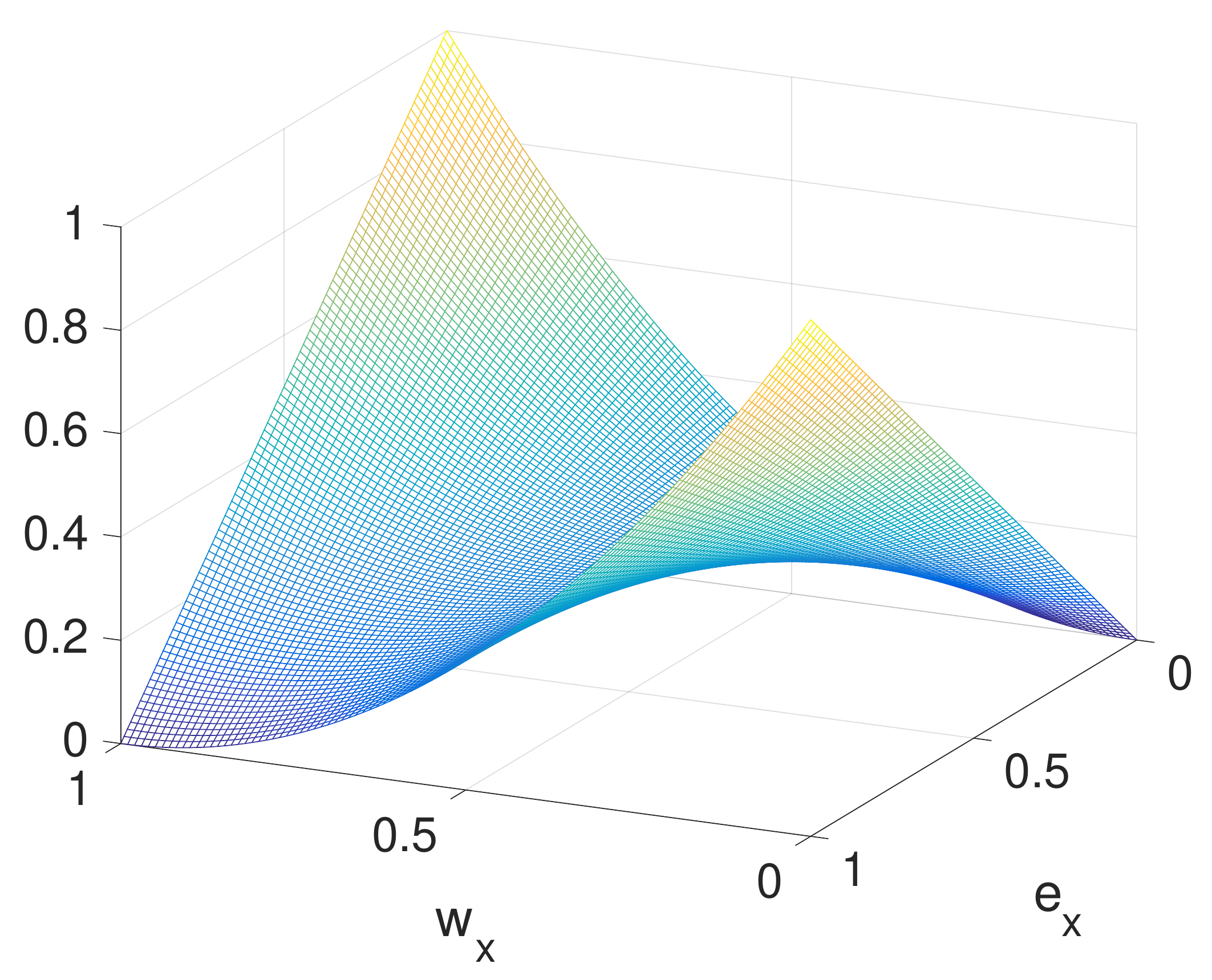}
    \caption{Loss function $\mathcal{L}_{\mathcal{S}}(e_{\mathbf{x}}, w_{\mathbf{x}})$ with $M=1$ without the $L_1$ regularizer. We bound $w_{\mathbf{x}}$ to be in $(0, 1)$ for our experiments.}
    \vspace{-.5em}
    \label{fig:loss_func}
\end{wrapfigure}

Valuating the exact significance of each training sample to maximize the final accuracy of the main network is computationally intractable~\cite{Graves2017}. The influence functions in~\cite{Koh2017} could be a potential solution to estimate the final accuracy with less computational burden but still requires significant computational cost at the initialization of every iteration of training the main network.
Instead of estimating the final accuracy, we propose to simplify the problem of sample-wise significance valuation to a local optimal policy that predicts the weights of training samples at the current iteration of the training.

%
Let $w_\mathbf{x}$ be a weight of training sample, $\mathbf{x}$, predicted by \snet, $\mathcal{S}$. 
Let $\mathcal{L}_{\mathcal{F}}(\mathcal{F}(\mathbf{x}), t_{\mathbf{x}})$ be an objective function for the main network $\mathcal{F}(\cdot)$ to compute an error between $\mathcal{F}(\mathbf{x})$ and its target label $t_\mathbf{x}$. We define an objective function that \snet\ minimizes, $\mathcal{L}_{\mathcal{S}}(\cdot)$, as follows:
\begin{equation}
\begin{split}
        \sum_{\mathbf{x} \in \mathbf{X}}^{} \left((1 - w_{\mathbf{x}})^2 e_{\mathbf{x}} +
               w_{\mathbf{x}}^2 \max(M - e_{\mathbf{x}}, 0) \right) 
               + \alpha \sum_{p \in \mathbf{W}_{\mathcal{S}}}^{} {\lVert p \rVert}_1,
\end{split}
\label{eq:Loss_S}
\end{equation}
where $w_{\mathbf{x}} = \mathcal{S}(\mathbf{x})$, $e_{\mathbf{x}} = \mathcal{L}_{\mathcal{F}}(\mathcal{F}(\mathbf{x}), t_{\mathbf{x}})$, $\mathbf{W}_{\mathcal{S}}$ is parameters of the \snet\ $\mathcal{S}$, and $\alpha$ is a balancing hyper-parameter for the regularizer of the \snet. $\mathbf{X}$ is a set of training images and $M$ is a margin hyper-parameter. 
We plot $\mathcal{L}_{\mathcal{S}}$ with $M=1$ except the $L_1$ regularizer in Figure~\ref{fig:loss_func}.

As shown in the Figure~\ref{fig:loss_func}, the objective function is a non-negative saddle like function with minima at $(w_x, e_x) = (0, 0)$ or $(1, 1)$ and with maxima at $(w_x, e_x) = (0, 1)$ or $(1, 0)$.
Thus, the $\mathcal{L}_{\mathcal{S}}$ promotes the sample weight to be high when the error (loss) of the sample by the main network is high and vice versa. Therefore, the \snet\ encourages to increase weight $w_x$ when error of the main network $e_x$ is high and promotes to decrease weight when the error is low.

Note that we bound the weight value by the \snet\ to be in $(0, 1)$ by a Sigmoid layer at the end of the \snet, since its multiplication to the gradient without the bound may cause overshooting or undershooting of the main network~\cite{Hinton2007}.

\begin{wrapfigure}{r}{.6\textwidth}
    \vspace{-1em}
    \centering
    \includegraphics[width=0.98\linewidth,bb=8 8 861 441]{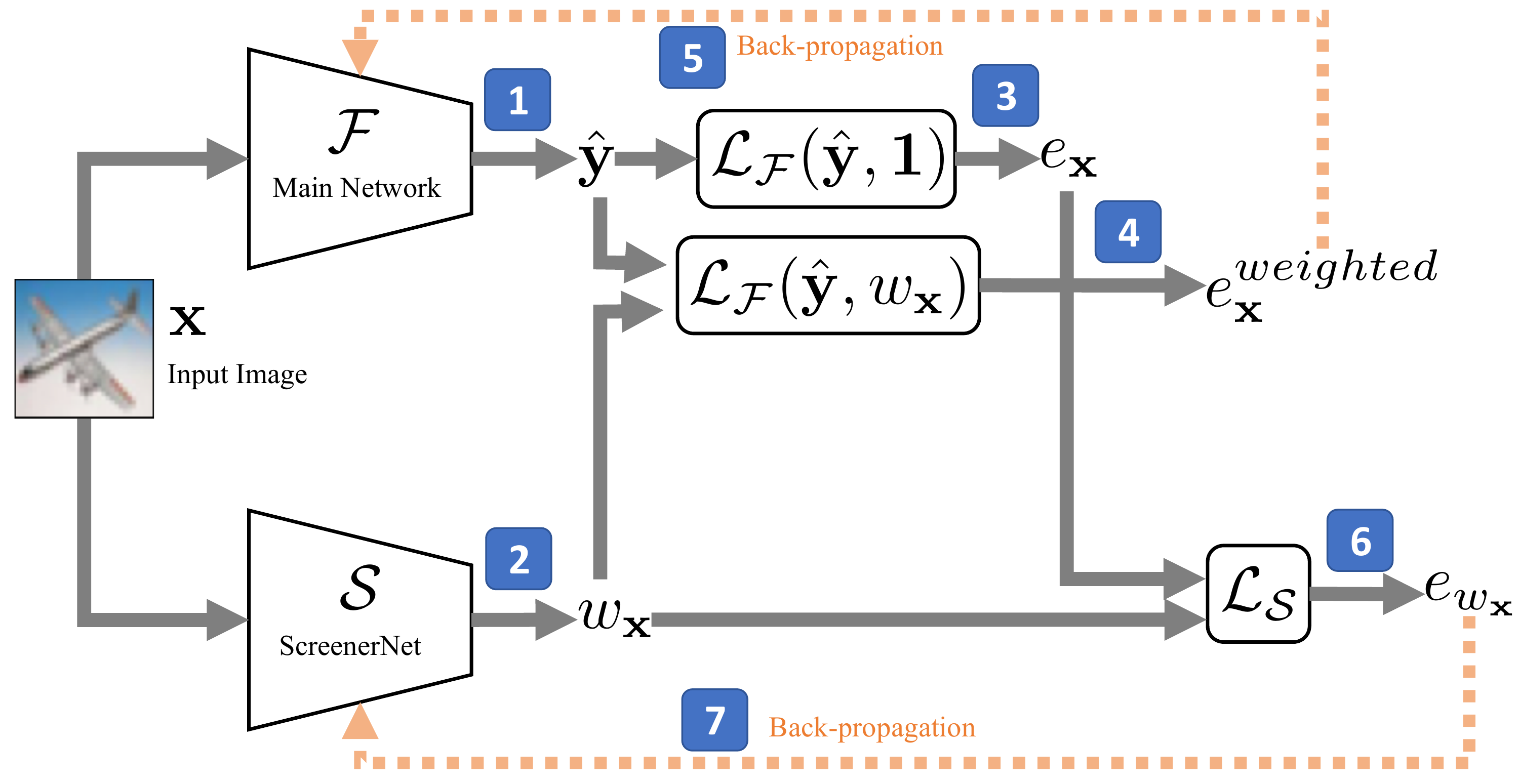}
    \caption{Optimization path of a \snet\ attached network. The numbers in blue boxes show the order of gradient update.}
    \vspace{-.5em}
    \label{fig:overview}
\end{wrapfigure}

\vspace{1em}\noindent\textbf{Optimization.}
Optimizing the loss of the \snet\ augmented network is not trivial as the loss is non convex. 
Thus, we employ the block-coordinate descent to optimize the network and depict the gradient path of the subset of variables in Figure~\ref{fig:overview}. The numbers in the blue boxes in the figure denotes the order of updating the gradient.

By the block coordinate descent, at each iteration of the epochs in the training phase, we predict the weight, $w_{\mathbf{x}}$ of the training sample $\mathbf{x}$ to update the main network $\mathcal{F}$ using the error, $e_{\mathbf{x}}$ from $\mathbf{x}$ and the weight $w_{\mathbf{x}}$. 
Then, we update \snet\ using the error, $e_{\mathbf{x}}$. 
The training procedure of \snet\ augmented deep network is summarized as Algorithm~\ref{algo:TrainSnet}.

    \begin{algorithm}[h]
        \DontPrintSemicolon
        \caption{Training a network with \snet}
        \label{algo:TrainSnet}
        \SetAlgoLined
        \emph{Given}: training samples $(\mathbf{X}, t_\mathbf{X})$,
        main network $\mathcal{F}$, \snet\ $\mathcal{S}$\;
        Initialize $\mathcal{F}$, $\mathcal{S}$\;
        \For{each iteration of the training}{
            $\mathbf{x} \gets$ Sample($\mathbf{X}$).    \tcc*{sample a mini-batch from ${\mathbf X}$}
            $w_\mathbf{x} \gets \mathcal{S}(\mathbf{x})$.    \tcc*{predict an weight of a sample}
            $\hat{\mathbf{y}} \gets \mathcal{F}(\mathbf{x})$.    \tcc*{prediction from the main network}
            $e_{\mathbf{x}}^{weighted} \gets \mathcal{L}_{\mathcal{F}}(\hat{\mathbf{y}}, w_{\mathbf{x}})$.    \tcc*{compute a weighted error}
            $e_{\mathbf{x}} \gets \mathcal{L}_{\mathcal{F}}(\hat{\mathbf{y}}, \mathbf{1})$. \tcc*{compute an error of the sample}
            UpdateNetwork($\mathcal{F}$, $e_{\mathbf{x}}^{w}$).    \tcc*{train main network}
            UpdateNetwork($\mathcal{S}$, $e_{\mathbf{x}}$).    \tcc*{train \snet}
        }
    \end{algorithm}
    \vspace{-1em}

\vspace{0.5em}\noindent\textbf{Architecture of \snet.}
A larger \snet\ than the main network would have larger capacity than the main network, then it may require more sample to reliably predict the significance of the samples.
A simple \snet\ would not predict the significance well and likely predicts it like a uniform distribution.
We empirically found that \snet\ whose architecture is similar to that of the main network or slightly simpler performs the best. 
Further, we also try weight sharing with the main network (refer to the parameter sharing argument in Section~\ref{sec:param_sharing}).

\subsection{Extending Stochastic Sampling Based Approaches}
\label{section:Extension}
Since the \snet\ is a weight regression network, it can extend the stochastic sampling approaches such as Prioritized Expereience Replay (PER)~\cite{Schaul2016}.
The extension has two benefits; first, it may further improve the convergence speed and the final accuracy than the single deployment of either method. Second, it reduces the computation for the \snet. 
For instance, if we combine the \snet\ with the PER.
The PER determines the probability of a training sample to be selected by
$
P(\mathbf{x}) = \frac{p_{\mathbf{x}}^\alpha}{\sum_{\tilde{\mathbf{x}}^{} \in \mathbf{X}} p_{\tilde{\mathbf{x}}}^\alpha},
\label{eq:sampling_probability}
$
where $p_{\mathbf{x}} > 0$ is the priority of each sample $\mathbf{x}$, and $\alpha$ controls how much prioritization is used. When $\alpha = 0$, it is equivalent to the uniform sampling. The priority is defined as:
$
p_{\mathbf{x}} = | e_{\mathbf{x}} | + \epsilon,
\label{eq:PER_priority}
$
where $\epsilon$ is a very small constant to prevent from assigning zero priority to $\mathbf{x}$.
To extend the PER or any other sampling based methods by the \snet, we predict weights of $\mathbf{x}$ that the PER or other sampling based methods select (refer to the Sec.~\ref{sec:snet_ext}).

%% file: exp.tex
\subsection{Datasets}
We have evaluated our algorithm on two tasks: a Cart-pole task using the deep Q-learning~\cite{Hasselt2015, Mnih2013}, which is one of the most popular tasks in deep reinforcement learning and a supervised visual recognition using three popular vision datasets; Pascal VOC 2012~\cite{pascal-voc-2012}, CIFAR10~\cite{CIFAR10} and MNIST~\cite{MNIST}.

For the Cart-pole experiment, we use Cart-pole-v0 in OpenAI Gym~\cite{OpenAIGym}, which gets 4-dimensional input of the state of the cart and pole, and outputs 2 discrete actions to move left or right.
Pascal VOC 2012 has $5,717$ images in training and $5,823$ images in validation set.
It is one of the most popular benchmark datasets along with the ImageNet.
CIFAR10 has $50,000$ images with size of $32{\times}32$ in training and $10,000$ images in testing set. 
MNIST dataset has $60,000$ images with size of $28{\times28}$ in training and $10,000$ images in testing set. 
Both MNIST and CIFAR10 are widely used for neural network training analyses.

\subsection{Experimental Set-up}
\label{exp:exp_setup}

We describe the details of the neural network architecture of the main network and the \snet\ used in our experiments on each dataset and their important hyper-parameters in the supplementary material.
We compare the \snet\ augmented network with baselines; the main network only and the state-of-the-art stochastic sampling based curriculum learning method, Prioritized Experience Replay (PER)~\cite{Schaul2016}, which is the only comparable to ours.
For the PER, we used importance-sampling weights $w_i = \left( \frac{1}{N}\cdot\frac{1}{P(i)} \right)^\beta$, where $\beta$ is linearly annealed from $0.4$ to $1.0$ until $40,000$ steps then is fixed to $1.0$. 
We set the discount parameter $\gamma$ to be $0.99$ and the replay memory to be a sliding window memory of size $50,000$. 
The PER processes mini-batches of $32$ samples (visual recognition task) or 32 transitions (deep Q-learning) sampled from the memory.

We compare the speed of convergence in training and the average reward per episode for the Cart-pole task or the final classification accuracy for the supervised visual recognition tasks (Pascal VOC 2012, CIFAR10 and MNIST).

\subsection{Faster and Better Convergence by \snet} 
\label{sec:learning_speed}

\begin{figure}[h]
\vspace{-.5em}
\centering
\subfloat[Cart-pole]{
\includegraphics[width=0.24\linewidth,bb=10 10 564 350]{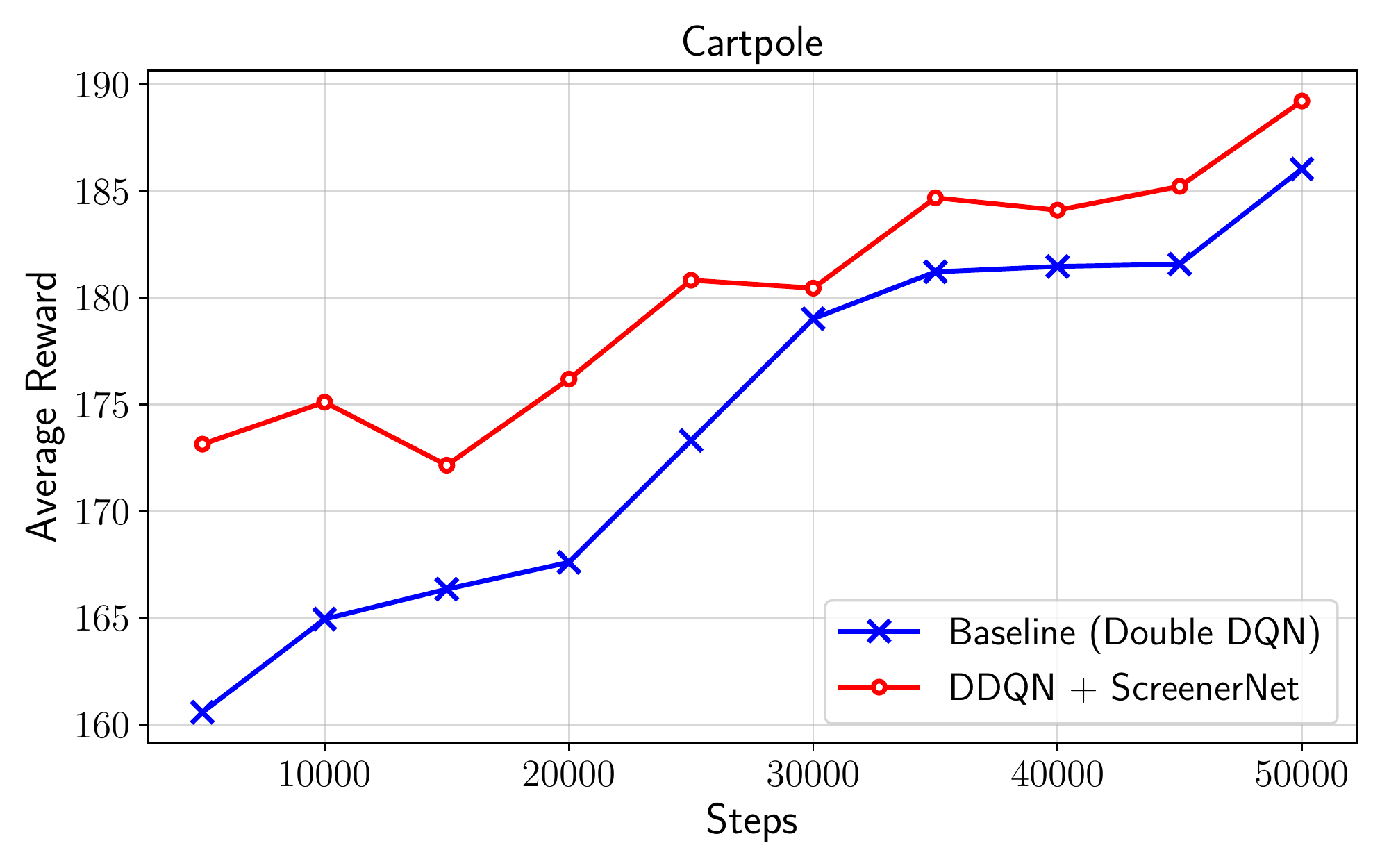}
}
\subfloat[Pascal VOC 2012]{
\includegraphics[width=0.24\linewidth,bb=10 10 565 350]{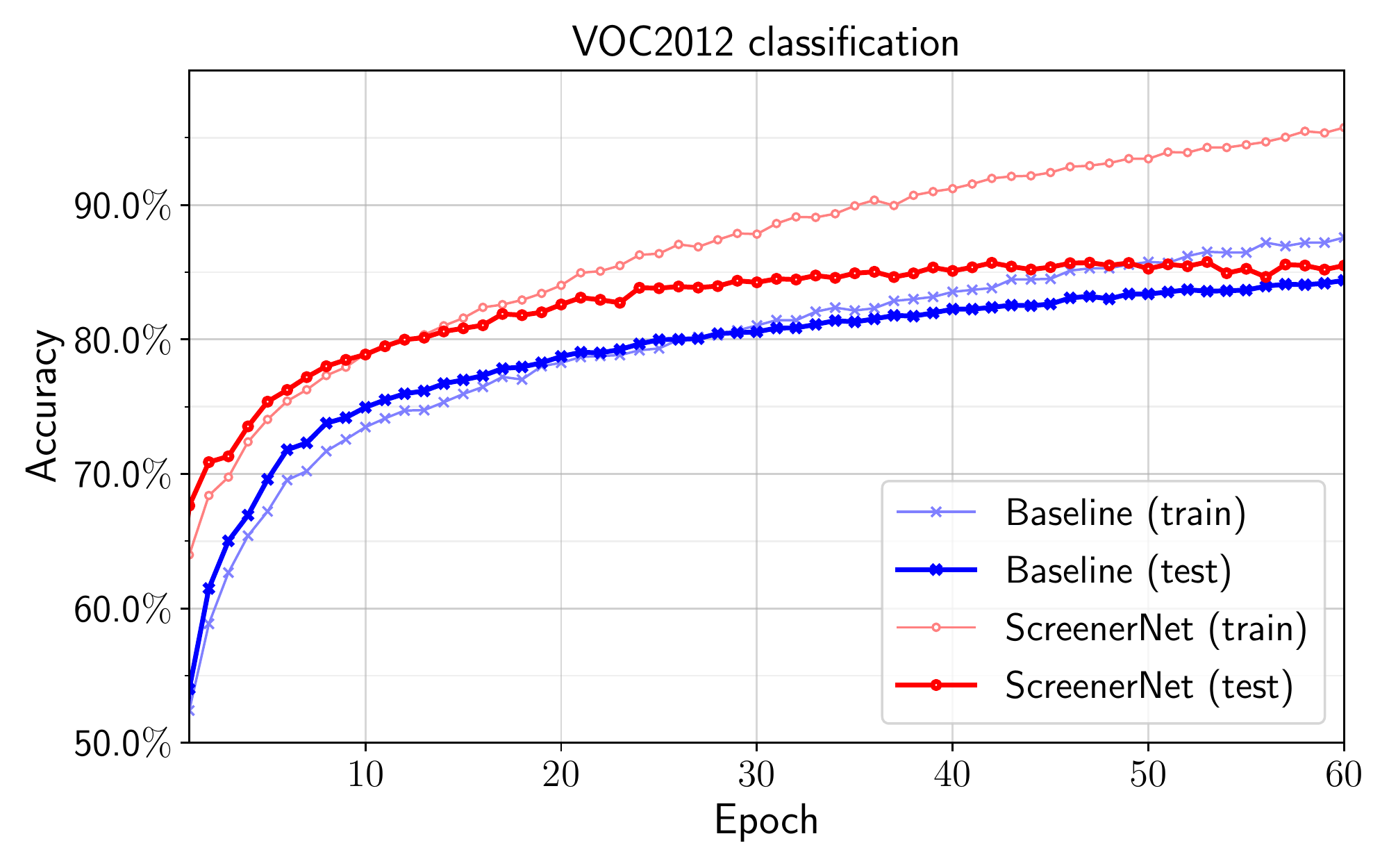}
}
\subfloat[CIFAR10]{
\includegraphics[width=0.24\textwidth,bb=10 10 565 350]{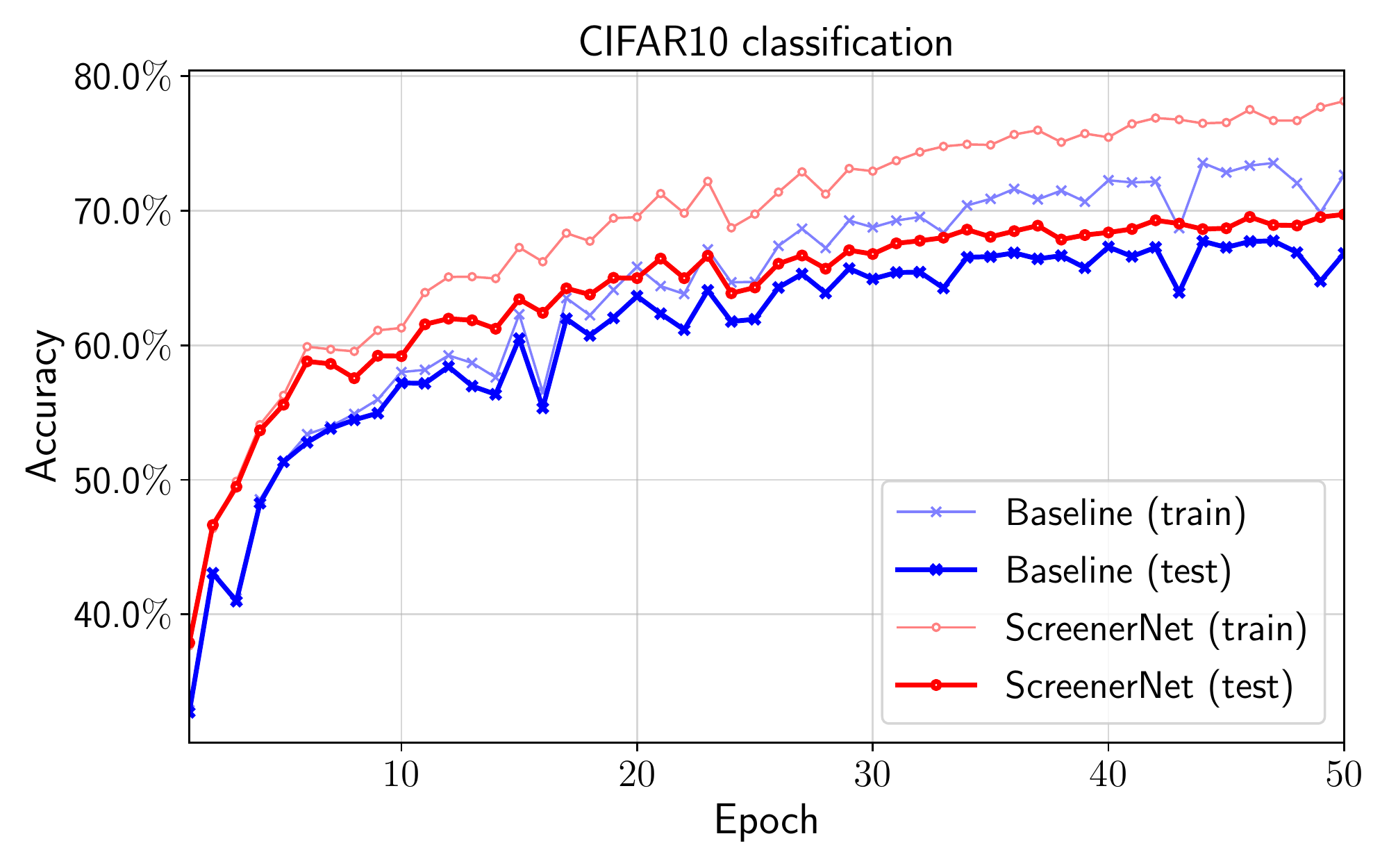}
}
\subfloat[MNIST]{
\includegraphics[width=0.24\linewidth,bb=10 10 565 350]{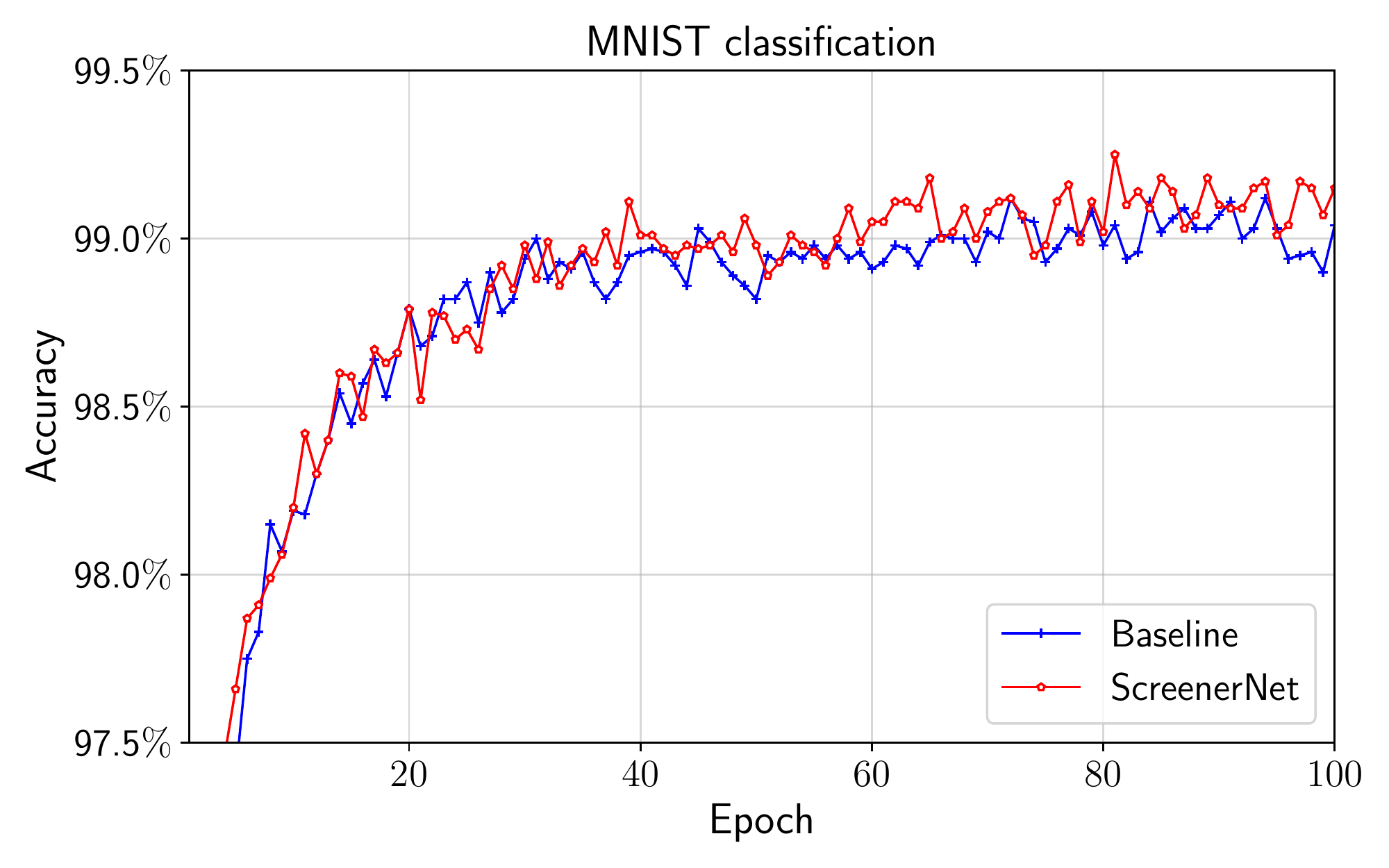}
}
\caption{Comparison of the test accuracy of main network ({\color{blue}Blue}) and \snet\ augmented network ({\color{red}Red})}
\vspace{-.5em}
\label{fig:snet_main_results}
\end{figure}

We compare the test accuracy curves of the baseline network (main network only) and the \snet\ augmented network (detnoted as `\snet') in Figure~\ref{fig:snet_main_results} for all four experiments.

In the Cart-pole task, the average reward obtained at every $5,000$ training step is shown in Figure~\ref{fig:snet_main_results}-(a). The \snet\ begins with higher average reward than the baseline Double DQN (DDQN) and clearly shows the higher overall gain comparing to the baseline during all epochs. 

For the supervised visual recognition tasks on three vision datasets, \snet\ also improves the convergence speed and the overall accuracy.
In the experiments with Pascal VOC 2012 dataset (Figure~\ref{fig:snet_main_results}-(b)), \snet\ helps the main network to improve the learning speed at early epochs even though VGG-19 is pre-trained with a much larger dataset (ImageNet). \snet\ also improves the accuracy when the networks are close to convergence though the gain diminishes.
We believe that the \snet\ helps better training rather than overfitting as the the sample significance that the \snet\ learns is quite different from the classification objective that the main network learns (refer to an empirical evidence for this argument in the discussion in Sec.~\ref{sec:param_sharing}). 
With CIFAR10 dataset shown in Figure~\ref{fig:snet_main_results}-(c), with the simple main network architecture, we observe that \snet\ yields improvements in the learning speed and final accuracy. 
With MINST dataset shown in Figure~\ref{fig:snet_main_results}-(d), \snet\ augmented networks converge slightly faster than the baseline and marginally improve the accuracy. It is because there is not much room to improve as the baseline already performs nearly perfect. MNIST and CIFAR10 experiments are mainly for the analysis purposes (Sec.~\ref{sec:qual_anal} and \ref{sec:param_sharing}).

In addition, it is of interest that when the network capacity varies, how much the \snet\ helps. We show effect of variously large \snet\ networks for the accuracy and the convergence speed in Section 2 in the supplementary material.

\begin{figure}[h]
\vspace{-.5em}
\centering
\subfloat[Cart-pole]{
\includegraphics[width=0.24\linewidth,bb=10 10 565 350]{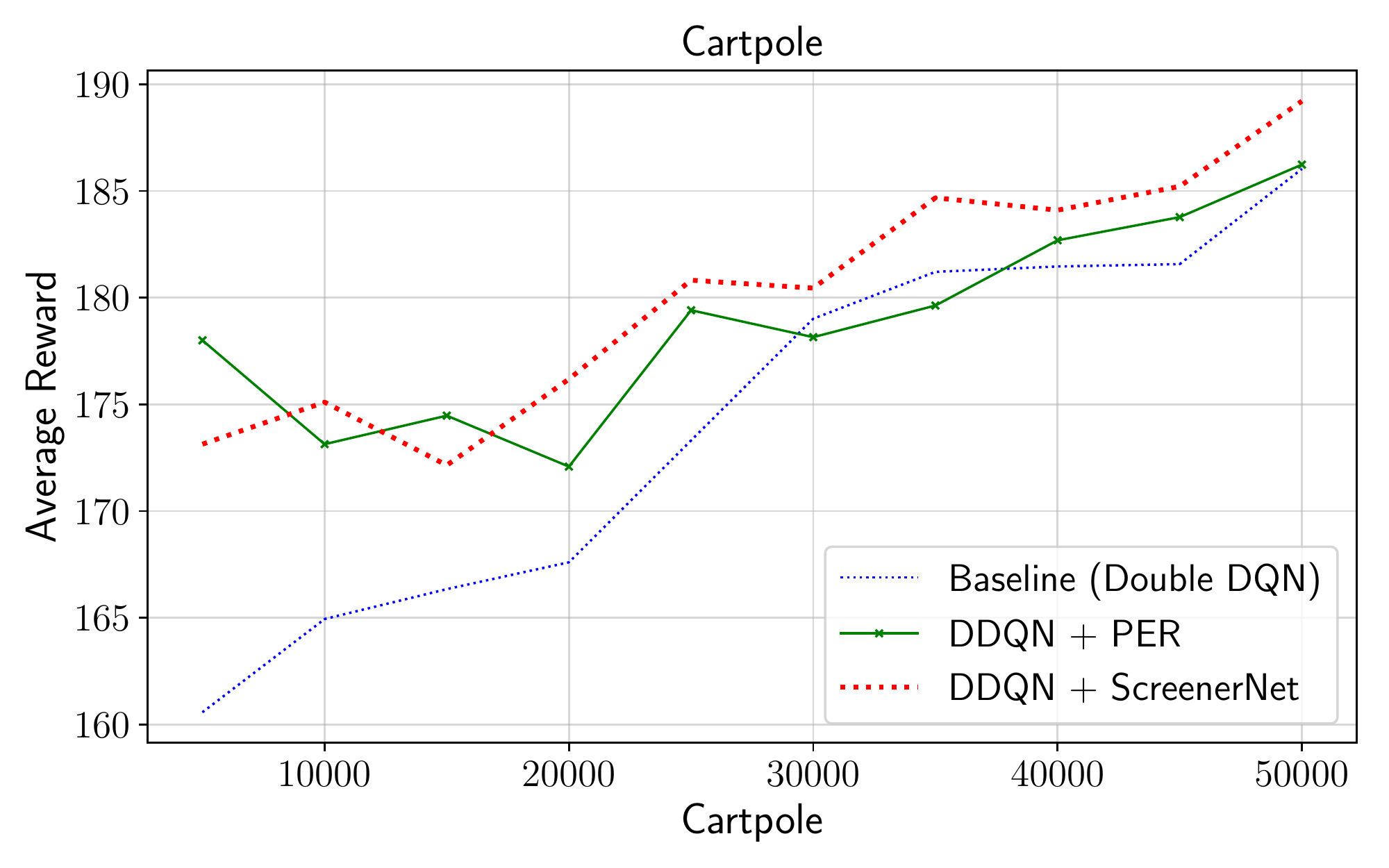}
}
\subfloat[Pascal VOC 2012]{
\includegraphics[width=0.24\textwidth,bb=10 10 565 350]{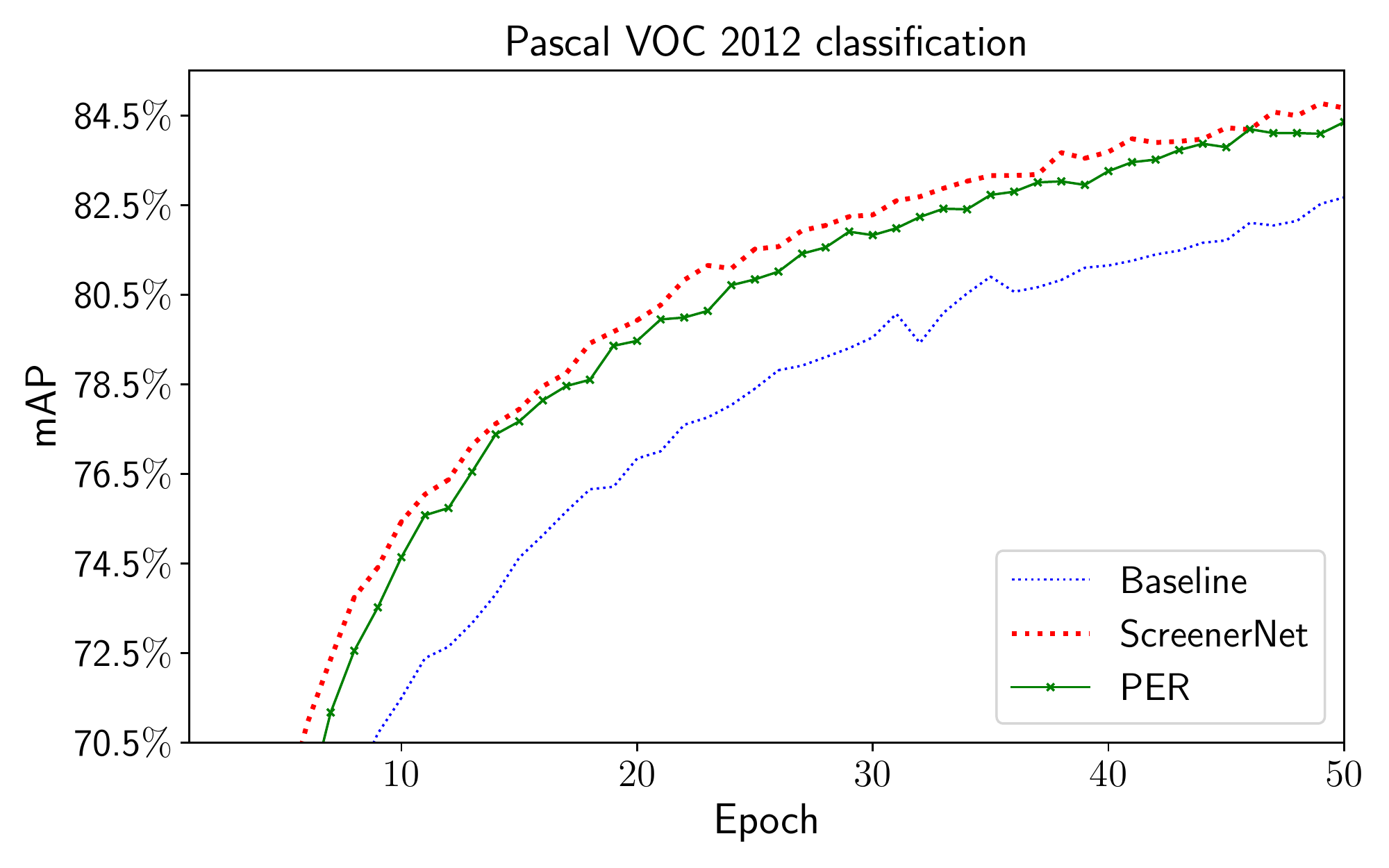}
}
\subfloat[CIFAR10]{
\includegraphics[width=0.24\textwidth,bb=10 10 565 350]{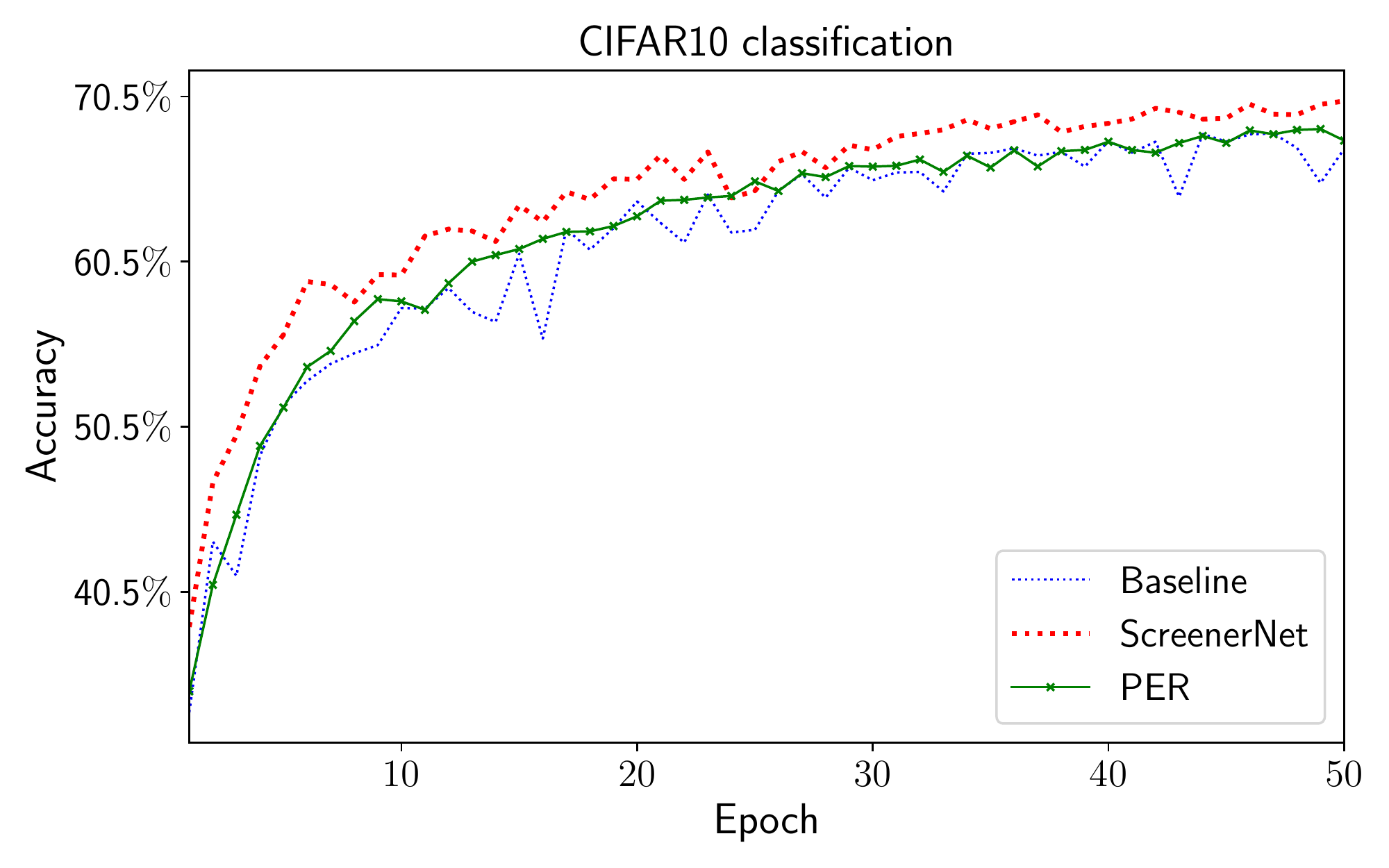}
}
\subfloat[MNIST]{
\includegraphics[width=0.24\textwidth,bb=10 10 565 350]{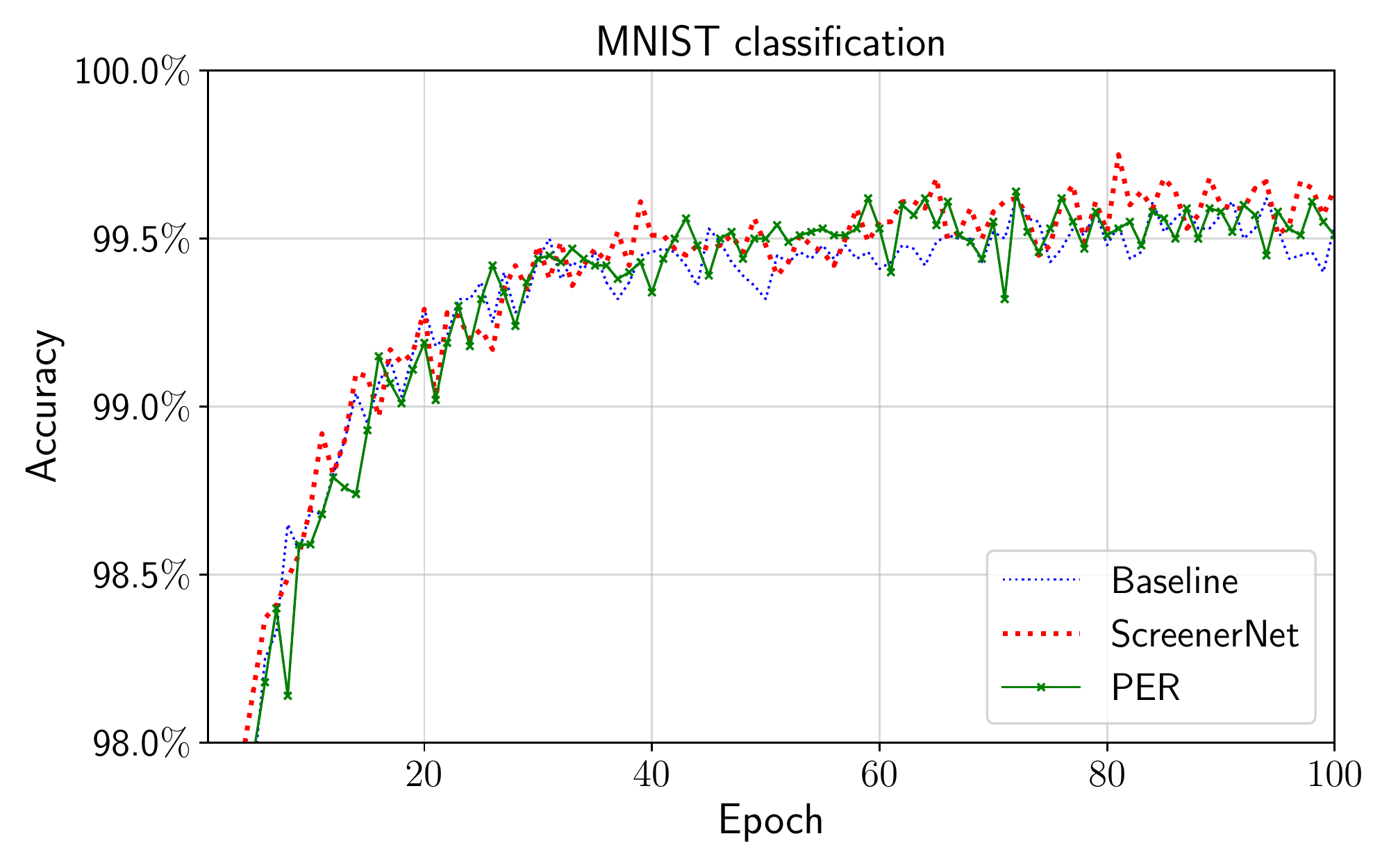}
}
\caption{Comparison with PER ({\color{green}Green}), Baseline (the main network only, {\color{blue}Blue}) and \snet\ augmented Network ({\color{red}Red})}
\label{fig:comparison_with_per}
\end{figure}

\vspace{0.5em}\noindent\textbf{Compare with Prioritized Experience Replay (PER).}
While the PER selects samples by a hard decision, \snet\ selects samples by a soft decision in a form of weight. 
They could be both comparable and able to be combined together, which we will discuss further in Sec.~\ref{sec:snet_ext}.
To see the benefit of \snet\ over the PER~\cite{Schaul2016}, we compare \snet\ with PER for all four datasets in Figure~\ref{fig:comparison_with_per}. We observe that \snet\ exhibits better learning curves of the main network than the PER in all four evaluations.

\subsection{Combined with Stochastic Sampling Methods for Deep Q-Learning}
\label{sec:snet_ext}

Since the stochastic sampling methods determine whether the samples to be included in the training or not by a defined probability, we can extend them by the \snet\ to investigate the synergistic effect of both methods.
Particularly, we can extend the \snet\ in two ways; first, \snet\ provides weights on the selected samples by the stochastic sampling methods. Second, the sampling selects the sample by the \snet's weight.
We evaluate both scenarios with the Prioritized Experience Replay (PER) as a choice of stochastic sampling methods for the deep Q-learning task using the cart-pole task as shown in Figure~\ref{fig:cart-pole_extension}.

\subsubsection{\snet\ Weighs Samples Selected by the Prioritized Experience Replay (PER)}

\begin{wrapfigure}{r}{.5\textwidth}
    \centering
    \vspace{-1em}
    \includegraphics[width=0.9\linewidth,bb=10 10 564 348]{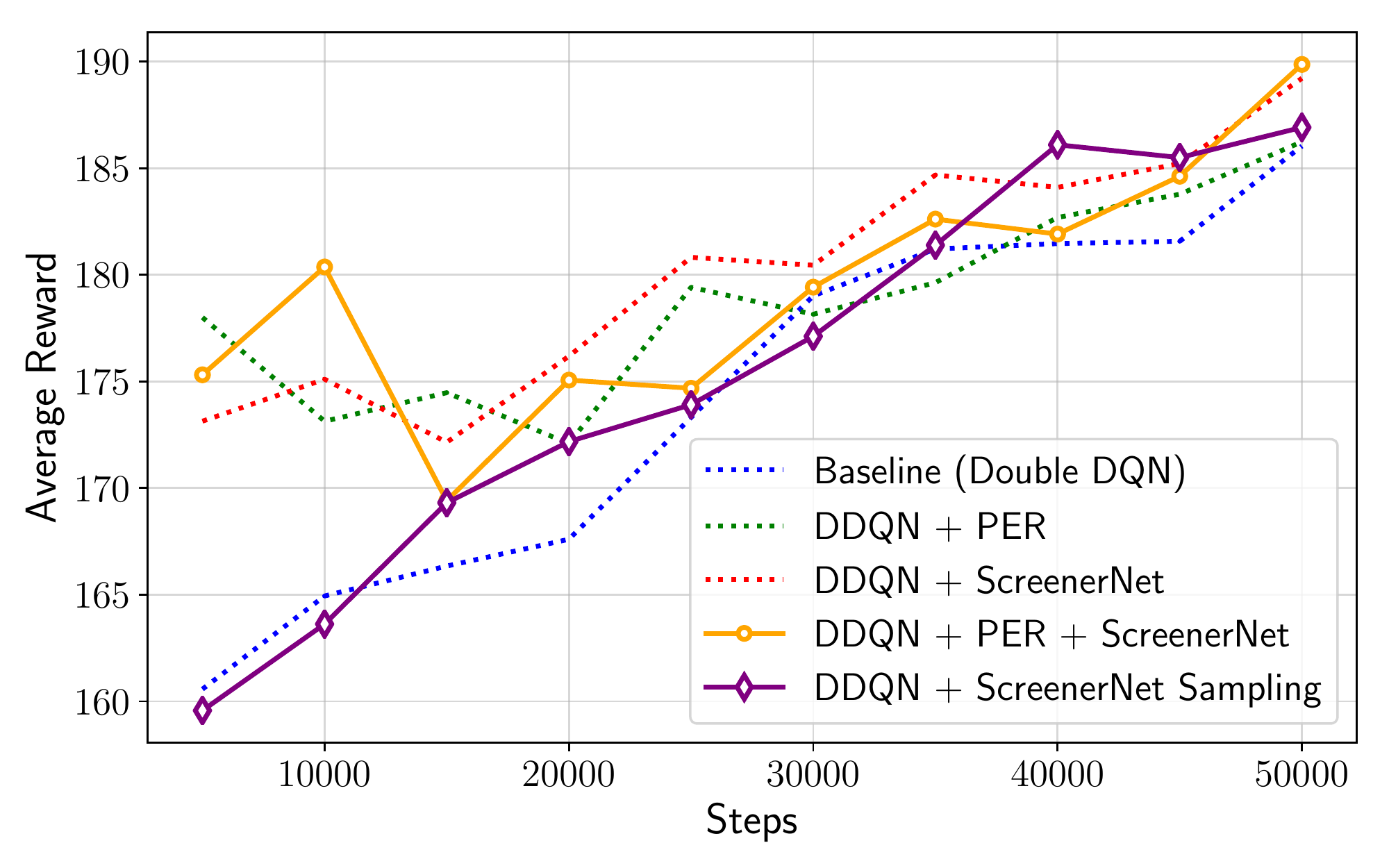}
    \caption{Learning curves for the cart-pole problem by average reward. Extension of combination of \snet\ and PER ({\color{orange}Orange}) and stochastic sampling using \snet\ ({\color{violet}Purple})}
    \vspace{-1em}
    \label{fig:cart-pole_extension}
\end{wrapfigure}

As a first scenario, we weigh the samples by the \snet\ on the samples selected by the PER (denoted as DDQN + PER + \snet\ ({\color{orange}orange solid line} in Figure~\ref{fig:cart-pole_extension}).
This extension outperforms the DDQN + PER ({\color{green}green dotted line}) but performs worse than the DDQN + \snet\ ({\color{red}red dotted}); since \snet\ weighs already selected samples, DDQN+PER+\snet\ performance could be bounded by the performance of the DDQN+\snet.

Interestingly, the \snet\, the PER and PER+\snet\ show similar learning progress; it begins with high accuracy but falls then gradually increases. 
Since they all have a sample weighting scheme, we believe that the methods with weighting schemes quickly learn the easy solutions in the beginning of learning (\emph{e.g.}, move the pole slightly from the initial position) then learn more difficult solutions thus the rewards drop in the early stage of learning then increase.

\subsubsection{PER with Probability by \snet\ }
In the PER, the probability that determines the priority of samples to be selected is defined by the error of the main network. For the second scenario of the \snet\ extension, we replace the probablity $p_{\mathbf{x}}$ in Eq.~\ref{eq:PER_priority} with the output of \snet\ ($p_{\mathbf{x}} = \mathcal{S}(\mathbf{x}) + \epsilon$) and setting $\alpha=1$ in Eq.~\ref{eq:sampling_probability}.
The result is shown in {\color{violet}purple solid line} (DDQN+\snet\ Sampling in Figure~\ref{fig:cart-pole_extension}).
Comparing it to the DDQN + PER ({\color{green}green dotted line}), the DDQN + PER outperforms the \snet\ Sampling in the beginning of training where \snet\ is not trained enough but eventually the \snet\ probability outperforms the direct error value.

\subsection{Qualitative Analysis}
\label{sec:qual_anal}

\begin{figure}[h]
\centering
\vspace{-1em}
\begin{minipage}{.475\textwidth}
    \centering
    \subfloat{
    \includegraphics[width=0.48\linewidth,bb=0 3 344 324]{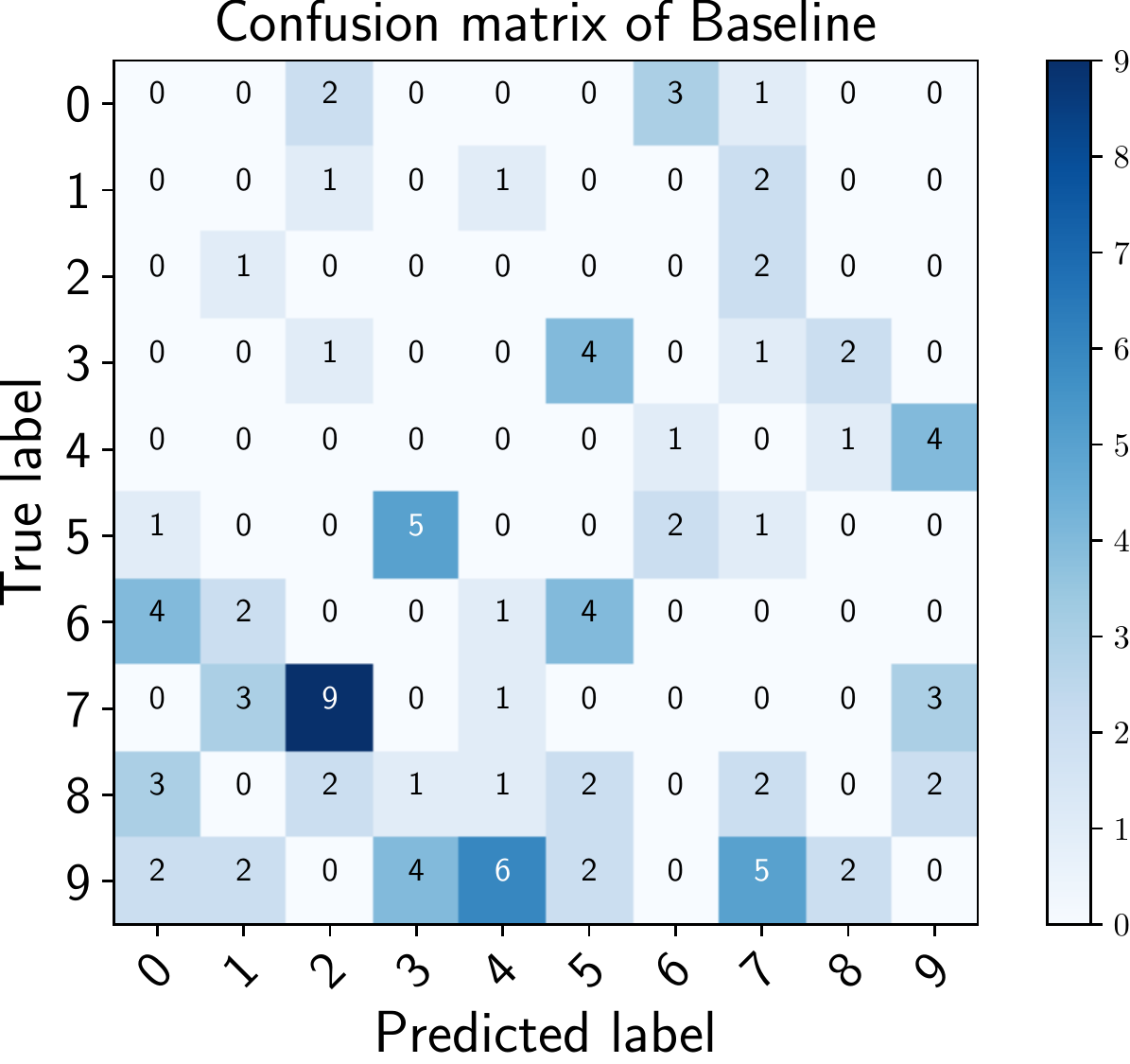}
    }
    \subfloat{
    \includegraphics[width=0.48\linewidth,bb=0 3 344 324]{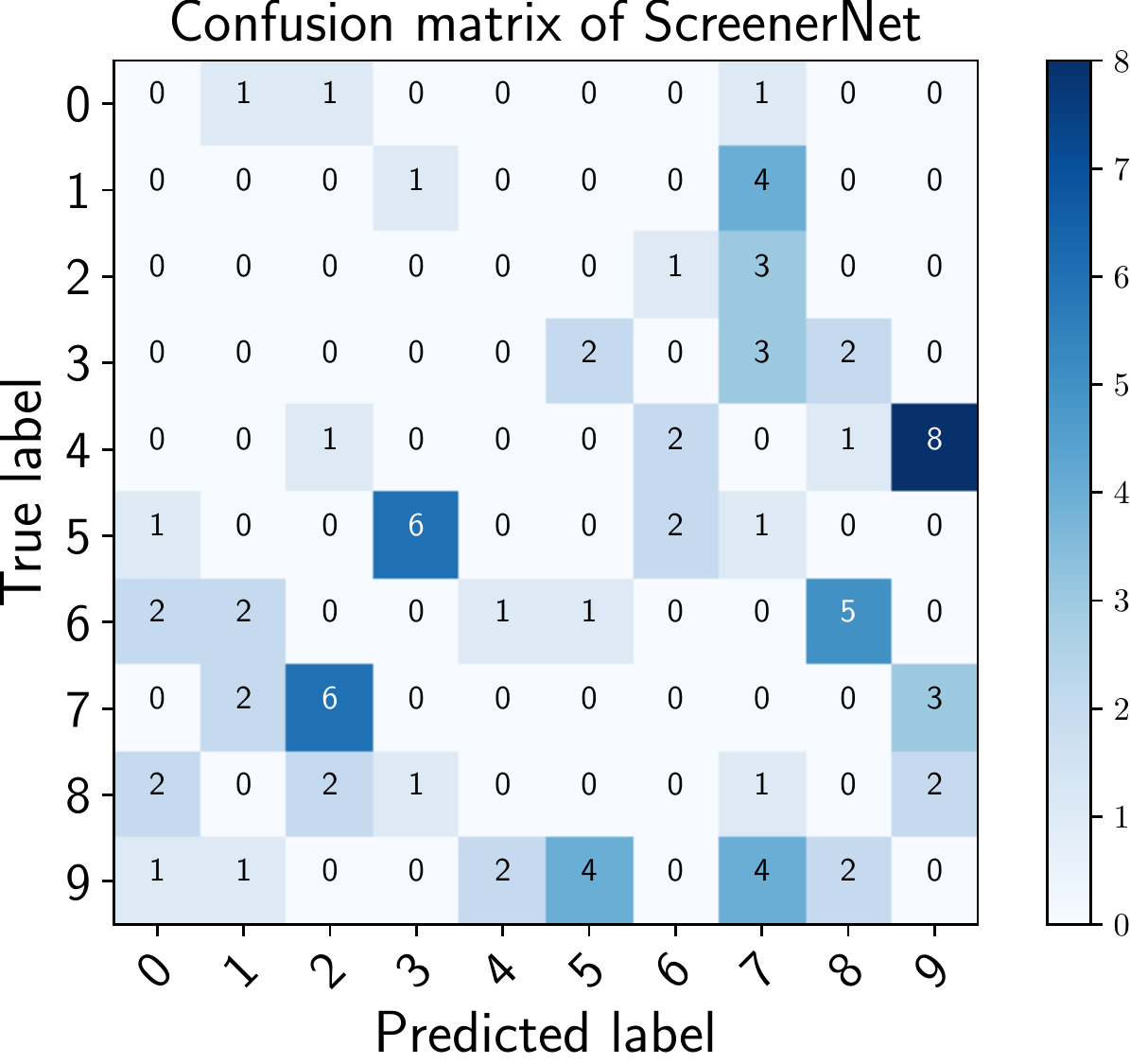}
    }
    \captionof{figure}{Confusion matrix \emph{only with the failed examples} for visual clarity. Each number in the cell shows the number of mis-classification. Note that the diagonal of the matrix is zero because successes are excluded for the visualization. \textbf{(Left)} by the network without \snet\, \textbf{(Right)} by \snet\ augmented network.
    }
    \label{fig:mnist_confusion_matrix}
    \vspace{-1em}
\end{minipage}
\hspace{1em}
\begin{minipage}{.475\textwidth}
    \centering
    \hspace{-.5em}
    \subfloat[Baseline only fails]{
    \includegraphics[width=0.585\linewidth,bb=3 3 214 126]{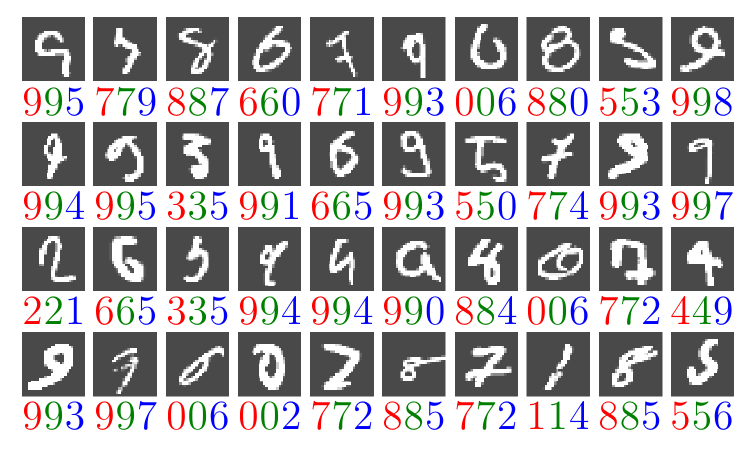}
    }
    \subfloat[\snet\ only fails]{
    \includegraphics[width=0.4\linewidth,bb=3 3 145 126]{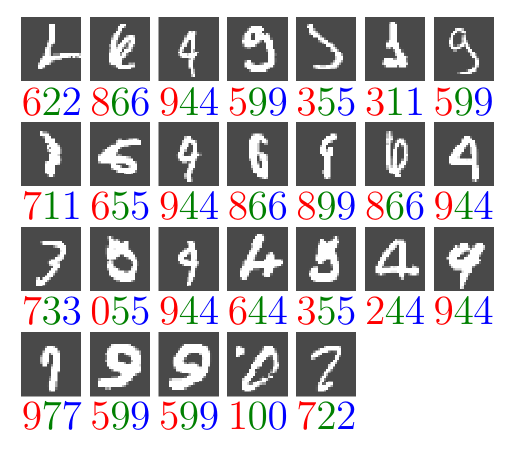}
    }
    \vspace{-.5em}
    \captionof{figure}{Comparison of exclusive failure examples. (a) baseline only fails but \snet\ succeeds and (b) \snet\ only fails but baseline succeeds. The three numbers under each sample image denote classification from \snet\ (in {\color{red}red}), ground-truth (in {\color{green}green}), and the baseline (in {\color{blue}blue}), respectively from left to right.}
    \vspace{-1em}
    \label{fig:quali_one_fail}
\end{minipage}
\end{figure}

\noindent\textbf{Error Analysis with MNIST.}
To better understand the effect of the \snet, we investigate failure cases of \snet\ augmented network (\snet) and the main network alone (baseline). We use MNIST dataset for the qualitative analses as it is better explanable than other image data or the Cart-pole task (Figure~\ref{fig:mnist_confusion_matrix} and \ref{fig:quali_one_fail}).

\begin{wrapfigure}{r}{.5\textwidth}
    \centering
    \vspace{-3em}
    \subfloat[Weight progress of MNIST]{
    \includegraphics[width=0.85\linewidth,bb=12 7 454 310]{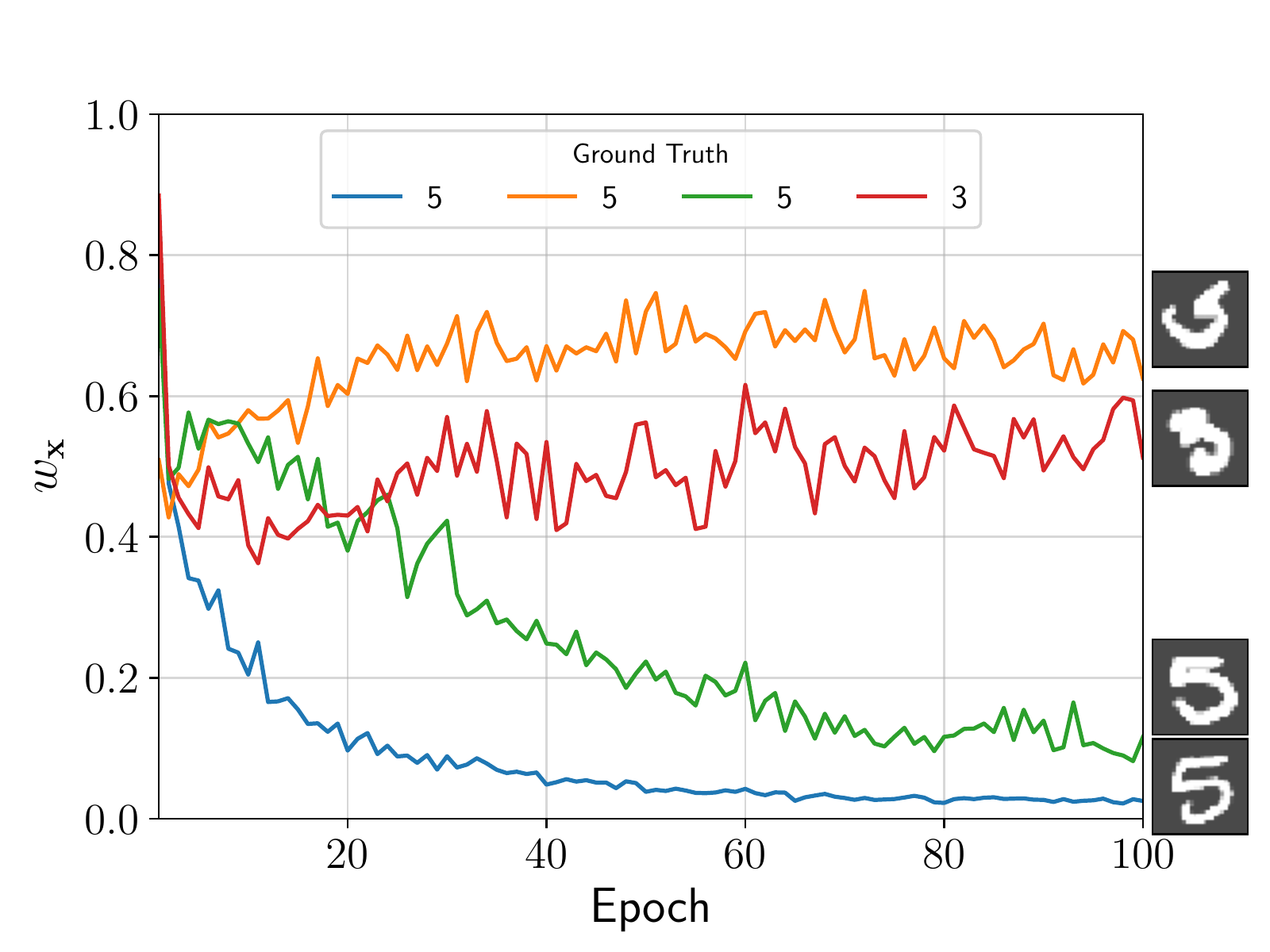}
    }\vspace{-1em}\\
    \subfloat[{\scriptsize Ends with highest weights}]{
    \includegraphics[width=0.45\linewidth,bb=3 1 116 38]{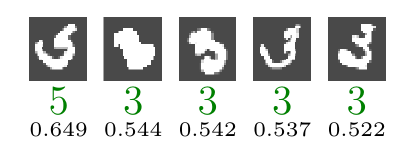}
    }
    \subfloat[{\scriptsize Ends with lowest weights}]{
    \includegraphics[width=0.45\linewidth,bb=3 1 116 38]{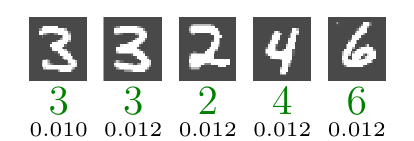}
    }
    \caption{Weight progress curves and their corresponding training images in MNIST. In (b) and (c) Green number denotes the ground truth label and the number below denotes the final weight value.}
    \vspace{-2em}
    \label{fig:weight_progress_mnist}
\end{wrapfigure}

\vspace{.5em}
We present confusion matrices of the baseline and \snet\ in Figure~\ref{fig:mnist_confusion_matrix} \emph{only with failed examples} for visual clarity. We also show qualtitative comparison that the failure cases that either only the \snet\ or the baseline classifies incorrectly in Figure~\ref{fig:quali_one_fail}.

It is observed that the widely spread confusions of the baseline are reduced overall by \snet\ in Figure~\ref{fig:mnist_confusion_matrix}.
But it also has a few new failures like mis-classification of $4$ into $9$. Note that the failure examples of $4$ into $9$ (Figure~\ref{fig:quali_one_fail}-(b)'s {\color{red}$9$}{\color{green}$4$}{\color{blue}$4$} examples) are extremely challenging.
Instead, \snet\ increases the precision of recognizing $4$, being more strict with classifying $1$, $7$, $8$, and $9$ into $4$. 
Similarly \snet\ increases recall for $8$ at the expense of precision.

\vspace{.5em}\noindent\textbf{Easy and Difficult Training Samples Selected by \snet.}
As the learning of \snet\ augmented neural network proceeds, difficult samples should receive high attention in the training procedure by high weight, while easy samples should receive low attention by low weight.
We plot the tracked weight of some easy and difficult samples as the learning proceeds in Figure~\ref{fig:weight_progress_mnist}-(a) and also present the samples with highest and lowest weights at the end of the training in Figure~\ref{fig:weight_progress_mnist}-(b) and (c).

The samples with high weights are visually difficult to distinguish, while the ones with low weights are visually distinctive to the other class thus training with these in the later epoch would not add much value to improve the accuracy.

\begin{wrapfigure}{l}{.55\textwidth}
    \centering
    \vspace{-1em}
        \includegraphics[width=0.9\linewidth,bb=10 10 565 350]{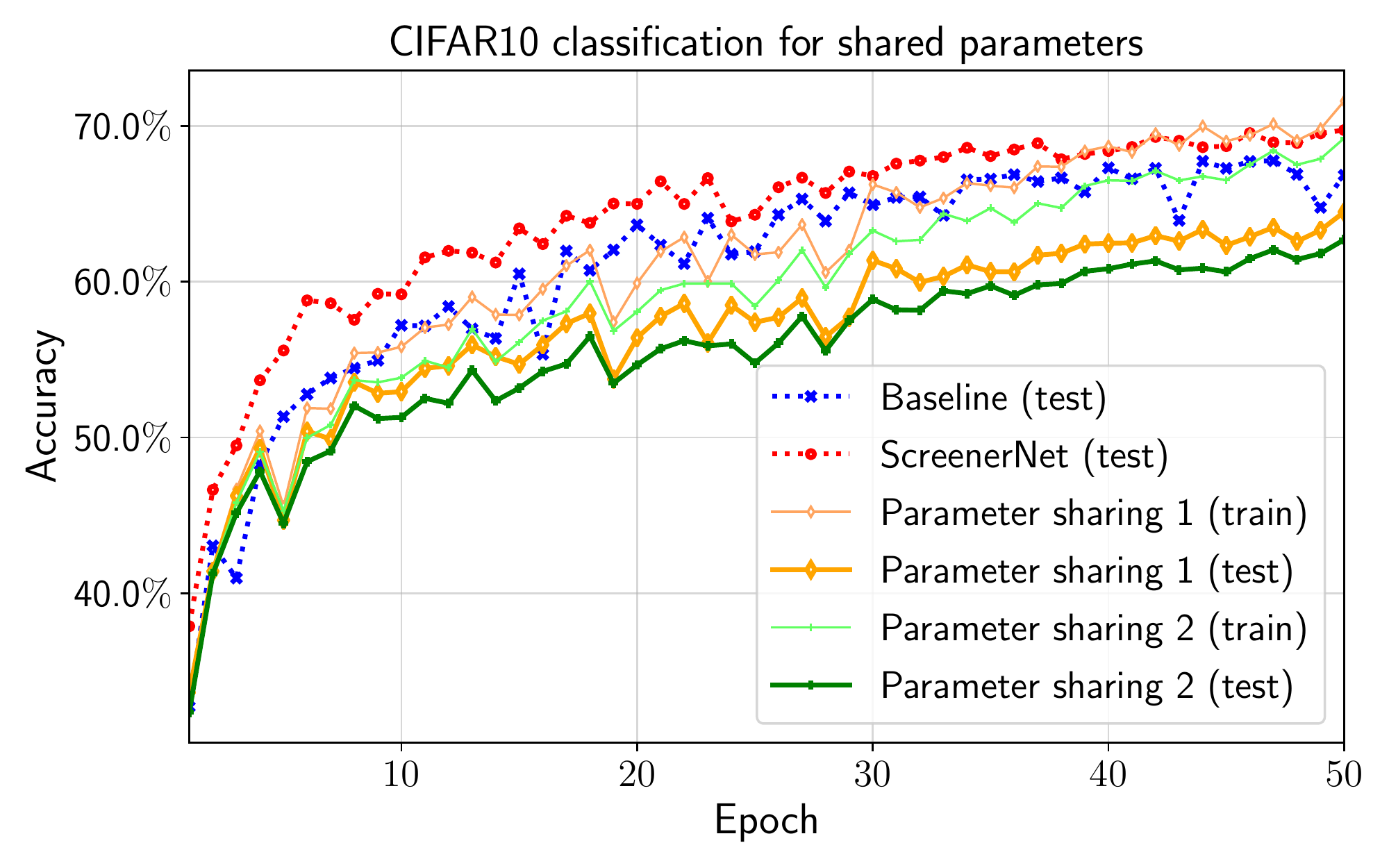}
    \caption{Accuracies of different \snet\ configurations (amount of parameter shared) on CIFAR10 dataset with small architecture for the main network.}
    \vspace{-1em}
    \label{fig:snet_variations}
\end{wrapfigure}

\subsection{Parameter Sharing}
\label{sec:param_sharing}
We tried the network parameter sharing between common layers of the main network and the \snet.
If the parameter sharing improves the accuracy, it implies that \snet\ simply adds more capacity to the main network and helps overfit to the data.

We conduct the experiments using CIFAR10 dataset since it has a good balance between the complexity and size.
There are two parameter sharing scenarios we explored; first, \snet\ does not change the parameter of shared layers but fine-tunes the last FC\_$1$ layer (sharing 1). Second, \snet\ also updates the parameter of the shared layers (sharing 2).
As illustrated in Figure~\ref{fig:snet_variations}, both parameter sharing scenarios decrease the accuracy. 
It implies that as the \snet\ does not learn the same objective that the main network learns, rather it learns the behavior of the main network; even the low level signal to the \snet\ is different from that of the main network. 
We argue that the \snet\ helps the main network to learn faster and better not helps it overfit.

%% file: conclusion.tex
We propose to estimate the significance of the training samples for effective curriculum learning by augmenting a deep neural network, called \snet, to the main network and jointly train them. We demonstrated that the \snet\ helps training deep neural networks both by fast and better convergence in various tasks including visual classification and deep reinforcement learning such as deep Q-learning. Moreover, the \snet\ not only outperforms the state-of-the-art sampling based curriculum learning method such as the Prioritized Experience Replay (PER) (Figure~\ref{fig:comparison_with_per}) but also can extend it for further improvement (Figure~\ref{fig:cart-pole_extension}).

Note that the learning objective of \snet\ is not the same as that of the main network. 
Instead, the \snet\ estimates the probability that the main network will correctly classify the given sample or not. The architecture of the \snet\ is simpler than the target network and thus can be trained ahead of it in terms of training maturity, which we empirically found to perform the best.

Since the \snet\ is not a memory-based model, it can also be considered as an error estimator of the current state of the main network. Thus, the \snet\ can estimate the sample confidence at inference time, which is particularly useful for the real world reinforcement learning system, similarly to the adaptive classifier proposed by~\cite{zhou2017}.

In contract to~\cite{jiangZLLL17}, the \snet\ possibly boosts weight values of mislabeled training samples which may perturb a decision boundary of the main network, since its objective function views training samples with large errors as significant ones to train the main network.

\vspace{.5em}\noindent\textbf{Future work.}
We can extend the \snet\ to be progressively expanding its complexity from a simple network to a complex one by adding new layers as the training progresses, similarly to~\cite{wangRH17}.
Since the newly added layers to the \snet\ might lead the learning to be unstable, we can use a momentum in updating the weight as $w_{\mathbf{x}} = \lambda \mathcal{S}_{old}(\mathbf{x}) + (1 - \lambda) \mathcal{S}_{new}(\mathbf{x})$, where $\mathcal{S}_{old}$ and $\mathcal{S}_{new}$ are respectively previous and current networks, and $\lambda$ is a hyper-parameter in $[0,1]$.